\documentclass[pdflatex,sn-mathphys-num]{sn-jnl}

\usepackage{graphicx}%
\usepackage{multirow}%
\usepackage{amsmath,amssymb,amsfonts}%
\usepackage{amsthm}%
\usepackage{mathrsfs}%
\usepackage[title]{appendix}%
\usepackage{xcolor}%
\usepackage{textcomp}%
\usepackage{manyfoot}%
\usepackage{booktabs}%
\usepackage{algorithm}%
\usepackage{algorithmicx}%
\usepackage{algpseudocode}%
\usepackage{listings}%

\usepackage{subcaption}

\theoremstyle{thmstyleone}%
\theoremstyle{thmstyletwo}%

\theoremstyle{thmstylethree}%

\begin{document}

\title[Regression-based Pelvic Pose Initialization for Fast and Robust 2D/3D Pelvis Registration]{Regression-based Pelvic Pose Initialization for Fast and Robust 2D/3D Pelvis Registration}


\author[1,2,3]{\fnm{Yehyun} \sur{Suh}}\email{yehyun.suh@vanderbilt.edu}

\author[4]{\fnm{J. Ryan} \sur{Martin}}\email{john.martin@vumc.org}

\author*[1,2,3]{\fnm{Daniel} \sur{Moyer}}\email{daniel.moyer@vanderbilt.edu}

\affil[1]{\orgdiv{Department of Computer Science}, \orgname{Vanderbilt University}, \orgaddress{\street{2201 West End Ave}, \city{Nashville}, \postcode{37235}, \state{TN}, \country{USA}}}

\affil[2]{\orgname{Vanderbilt Institute of Surgery and Engineering}, \orgaddress{\street{2201 West End Ave}, \city{Nashville}, \postcode{37235}, \state{TN}, \country{USA}}}

\affil[3]{\orgname{Vanderbilt Lab for Immersive AI Translation}, \orgaddress{\street{2201 West End Ave}, \city{Nashville}, \postcode{37235}, \state{TN}, \country{USA}}}

\affil[4]{\orgdiv{Department of Orthopaedic Surgery}, \orgname{Vanderbilt University Medical Center}, \orgaddress{\street{1211 Medical Center Dr}, 
\city{Nashville}, \postcode{37232}, \state{TN}, \country{USA}}}



\abstract{This paper presents an approach for improving 2D/3D pelvis registration in optimization-based pose estimators using a learned initialization function. Current methods often fail to converge to the optimal solution when initialized naively. We find that even a coarse initializer greatly improves pose estimator accuracy, and improves overall computational efficiency. This approach proves to be effective also in challenging cases under more extreme pose variation. Experimental validation demonstrates that our method consistently achieves robust and accurate registration, enhancing the reliability of 2D/3D registration for clinical applications. Code is available at \url{https://github.com/yehyunsuh/Regression-Initialized-2D-3D-Pelvis-Registration}.}




\maketitle

\section{Introduction}
\label{sec:intro}
Accurate pelvic pose estimation is crucial for pre-operative planning, intra-operative navigation, and post-operative assessment in hip and lower (lumbar) spine surgeries \cite{fischer2020preoperative,inaba2016preoperative}.
While 3D imaging would be preferable from a diagnostic perspective, intra-operative fluoroscopy (2D) is more common, faster, and has a lower radiation dosage. Pre-/post-operative imaging often collect 2D radiography for similar reasons: availability, cost, dosage, and speed.
Thus, it is important to be able to accurately reconstruct pelvic pose from 2D imaging.




A number of pelvis-specific 2D/3D pose estimators have been proposed in the literature \cite{uneri20133d,gao2020ProST,gao2023ProST}. While these methods appear to have fidelity for a subset of cases, by their own reporting they fail in a non-trivial number of cases (at least $>30\%$ in \cite{gao2023ProST}). Surprisingly, we find that this is in part due to fragility of the methods with respect to initialization. Further, we find that a relatively simple learned initialization step significantly improves performance for all tested optimization methods, and reduces the number of optimization steps required, reducing effective run time by a large margin. While this learned initializer cannot itself directly predict pose parameters, it provides a ``close-enough'' starting point, which we show to improve later optimization-based estimator accuracy in the vast majority of cases.

In this paper, we propose a method that provides robust initialization for pelvic 3D pose estimation from 2D imaging. The proposed method is simple yet effective, and takes a step towards the consistency required for clinical viability, improving and complementing existing pose-estimator methods.

\section{Related Work}
Traditional approaches to 2D/3D registration can be broadly categorized into image-based and feature-based methods. Image-based methods achieve alignment by directly comparing the intensity values of 2D radiographs and digitally reconstructed radiographs (DRRs) from 3D models. They avoid the need for specialized equipment and can be easily incorporated into existing medical imaging and computational analysis pipelines \cite{chen2024fully}. Feature-based approaches focus on extracting and matching anatomical landmarks or geometric structures between 2D and 3D data. These methods tend to be faster and have a wider capture range, allowing alignment to be focused on selected subsets of the data \cite{guan2018review}.

Recent techniques implement embedding-based registration and direct pose regression methods. Embedding-based approaches map 2D and 3D data into a common feature space using deep learning models, optimizing the embedded similarity for more effective alignment. They enhance the ability of registration by effectively capturing complex relationships between 2D and 3D data \cite{gao2023ProST,gao2020ProST}. Direct pose regression methods have been developed to predict pose or the transformation parameters directly from images, reducing computational complexity compared to traditional iterative approaches, making them more efficient for clinical applications.

\section{Method}

\subsection{Preliminaries and Optimization-based Pose Estimation}

We represent rigid transformations as a rotation $R\in\text{SO}(3)$ and translation vector $t\in\mathbb{R}^3$. This transformation is often expressed as a transformation matrix for homogeneous coordinates:
\begin{equation}
T = \begin{bmatrix} R & t \\ 0 & 1 \end{bmatrix}, 
\end{equation}
We parameterize rotations $R$ using Euler angles $(r_x,r_y,r_z)$, which are rotations around the $x$, $y$, and $z$-axes. We parameterize translations $t$ represented by $(t_x,t_y,t_z)$ as displacements along those axes. These together form $\theta$, which parameterizes $\text{SE}(3)$.

\begin{figure}
    \centering
    \includegraphics[width=1\textwidth]{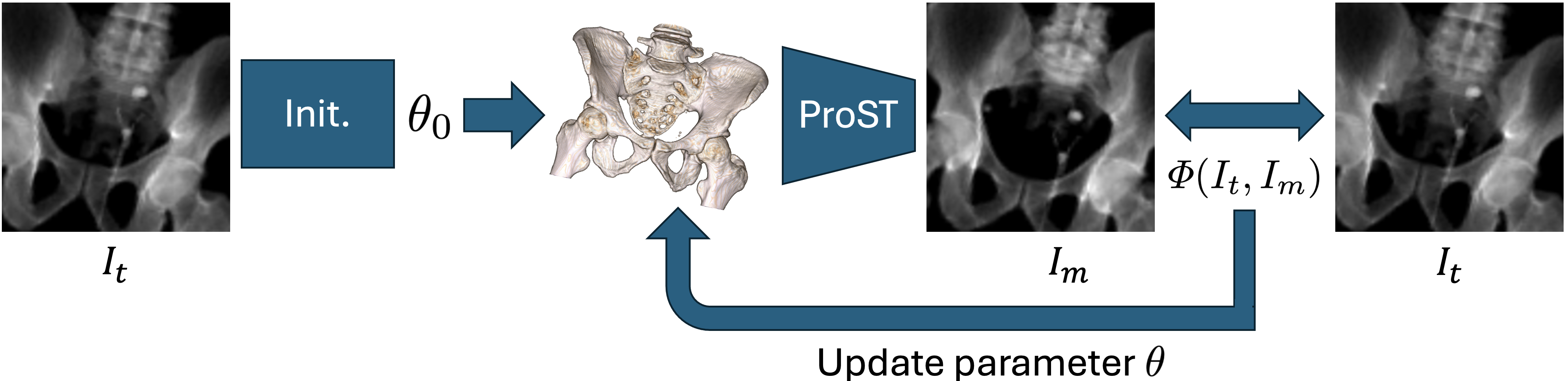}
    \caption{This figure diagrams the iterative registration framework; our initializer is at left, agnostic to the optimization framework at center and right.}
    \label{fig:method}
\end{figure}

Pose estimation methods in general aim to find either the subject/patient pose or the camera pose from a single 2D projection. For consistency, pose will refer to camera pose, though in our application these two cases are actually the same; radiographs rarely have meaningful backgrounds, and pelvic motion is assumed to be rigid (an element $\text{SE}(3)$ group). Pose estimation for pelvic imaging is thus a search over $\theta$ rigid transformations, looking for the rigid transformation $\theta^*$ that best aligns the projected 3D volume with a given 2D target image. 

Given a 3D image $V$, the observed 2D image $I_t$, a projection operator $\mathcal{P}$, and a particular $\theta$, the moving image $I_m(\theta)$ is obtained by applying rigid transformation $T(\theta)$ to $V$:
\begin{equation} 
I_m(\theta) = \mathcal{P}(T(\theta) V). 
\end{equation}
Optimization-based pose estimation methods choose an image similarity metric $\Phi$ and assume that $\theta^*$ is such that the similarity between $I_m$ and $I_t$ is maximized:
\begin{equation} 
\max_{\theta} \Phi(I_t, I_m(\theta)) = \Phi(I_t, \mathcal{P}(T(\theta^*) V)).
\end{equation}
The similarity measurement may be a standard image similarity metric \cite{penney1998gradncc} or might be a Euclidean distance in a learned feature embedding of the images. To determine the optimal pose, the optimization-based methods perform their namesake optimization, minimizing a loss function based on the similarity calculation:
\begin{equation} 
\theta^* = \arg\min_{\theta} \mathcal{L}(\theta). \label{opt}
\end{equation}
The optimization process is carried out iteratively using gradient-based methods to refine $\theta$ until convergence. 

\subsection{Regression and Registration}
Like many difficult optimization problems, for reasonable choices of $\Phi$ Eq \ref{opt} has numerous local minima. Standard local optimization method (e.g., gradient descent) are dependent on initialization $\theta_0$, and choices of $\theta_0$ far away from $\theta^*$ cause high error rates. However, even though estimating $\theta$ directly from the image also leads to poor solutions (Table \ref{tab:average}, first rows), using those first image-regression estimates serves as a strong initialization. As we demonstrate in Section \ref{sec:results}, this simple fact greatly improves results.

Our proposed method's procedure has the following steps (shown graphically in Fig \ref{fig:method}):
\begin{enumerate}
\item \label{reg-step-1} Given a target image $I_t$, estimate the initial pose parameters $\theta_0$ using an Initializer (Init.).
\item \label{reg-step-2} Apply the estimated pose $\theta_0$ to the pelvis CT image and generate a DRR using the Projective Spatial Transformer (ProST) \cite{gao2020ProST,jaderberg2015STN} or another differentiable DRR method.
\item \label{reg-step-3} Compute the alignment error between $I_m$ and $I_t$ using a similarity metric $\Phi$.
\item \label{reg-step-4} Update the pose parameter $\theta$ to minimize the error and refine the alignment.
\item Repeat from Step \ref{reg-step-2} until convergence.
\end{enumerate}

The Initializer function is defined ambiguously, and could generally be any function estimator taking an image to a valid $\theta_0$. We test three variants of a simple feed forward architecture, each with different input structures:
\begin{itemize}
\item \textbf{(Proposed,1)}, which takes only the target image $I_t$ as input.
\item \textbf{(Proposed,2,PE)}, which takes both the moving image $I_m$ and target image $I_t$ as input, as well as an image positional encoding \cite{dosovitskiy2021VIT} for each pixel concatenated as additional channels.
\item \textbf{(Proposed,2,PE,AC)} which includes the inputs to the \textbf{(Proposed,2,PE)}, plus the absolute coordinates of each pixel.
\end{itemize}

For each of these input signatures we use the same ResNet-18 architecture \cite{he2016resnet}, initialized with pre-trained weights from ImageNet \cite{deng2009imagenet} and a new first layer (to account for the varying number of input channels, none of which are RGB color channels unlike the original ImageNet domain). The output is $\theta_0$ directly, using global pooling and a final fully connected layer. All variants are trained using an MSE loss function on simulated poses of CT volumes and were trained for 1000 epochs.

While this initializer method adds a non-zero number of parameters to the overall pose-estimation methods, the runtime itself is faster than a single iteration of any of the pose-estimator methods.

\begin{table}[ht]
\centering
\caption{Quantitative comparison of registration methods with and without the proposed initializer. Column values, left to right: mean number of iterations required for convergence (Mean Iters), root mean square error (RMSE) and mean absolute error (MAE) for both rotation and translation, with translation errors further divided into in-plane (xy-Trans.) and out-of-plane (z-Trans.).}
\label{tab:average}
\begin{tabular}{lccccccc}
\hline
Method & Mean & Rot. & Trans. & Rot. & xy-Trans.& z-Trans.\\
(Init. Method) & Iters & RMSE & RMSE & MAE & MAE & MAE\\ \hline
\multicolumn{7}{l}{\textbf{Gao et al. \cite{gao2020ProST} Pose Parameter Range}} \\ \hline
Initializer (Original)       & - & 11.95 & 17.18 & 10.71 & 15.45 & 15.38\\
Initializer (Proposed,1)       & - & 6.12 & 15.93 & 5.26 & 9.70 & 20.92\\
Initializer (Proposed,2,PE)    & - & 8.38 & 20.00 & 7.24 & 14.52 & 23.86\\
Initializer (Proposed,2,PE,AC) & - & 7.29 & 20.33 & 6.26 & 11.30 & 28.95\\
\hline
Intensity \cite{gao2020ProST} (Original)         & 211.04 & 9.41 & 15.48 & 8.37 & 11.65 & 16.98\\
Intensity \cite{gao2020ProST} (Proposed,1)       & \textbf{110.56} & 3.48 & 9.99 & 2.98 & \textbf{5.54} & 13.88\\
Intensity \cite{gao2020ProST} (Proposed,2,PE)    & 131.42 & 5.48 & 14.29 & 4.70 & 10.27 & 17.24\\
Intensity \cite{gao2020ProST} (Proposed,2,PE,AC) & 121.07 & 4.57 & 12.98 & 3.87 & 7.15 & 18.80\\
\hline
MICCAI \cite{gao2020ProST} (Original)          & 263.13 & 12.41 & 35.81 & 11.03 & 31.18& 30.52\\
MICCAI \cite{gao2020ProST} (Proposed,1)        & 141.11 & \textbf{2.51} & \textbf{8.14} & \textbf{2.04} & 5.83 & \textbf{8.43}\\
MICCAI \cite{gao2020ProST} (Proposed,2,PE)     & 154.30 & 3.19 & 8.59 & 2.66 & 6.28& 9.29\\
MICCAI \cite{gao2020ProST} (Proposed,2,PE,AC)  & 154.91 & 3.02 & 11.57 & 2.50 & 7.21& 14.46\\
\hline
TMI \cite{gao2023ProST} (Original)          & 280.19 & 16.61 & 49.71 & 14.73 & 49.57 & 32.92\\
TMI \cite{gao2023ProST} (Proposed,1)        & 249.00 & 9.39 & 41.85 & 7.98 & 40.20 & 31.84\\
TMI \cite{gao2023ProST} (Proposed,2,PE)     & 253.76 & 8.85 & 42.59 & 7.51 & 38.24 & 38.64\\
TMI \cite{gao2023ProST} (Proposed,2,PE,AC)  & 252.87 & 8.74 & 45.86 & 7.46 & 38.67 & 47.55\\
\hline
\multicolumn{7}{l}{\textbf{Extended Pose Parameter Range}} \\ \hline
Initializer (Original)       & - & 24.20 & 28.61 & 21.96 & 26.08 & 25.71\\
Initializer (Proposed,1)       & - & 8.45 & 23.88 & 7.35 & 17.46 & 27.26\\
Initializer (Proposed,2,PE)    & - & 11.19 & 25.42 & 9.60 & 19.66 & 27.45\\
Initializer (Proposed,2,PE,AC) & - & 9.65 & 24.25 & 8.35 & 15.95 & 31.02\\
\hline
Intensity \cite{gao2020ProST} (Original)         & 247.77 & 23.85 & 30.98 & 21.29 & 25.21 & 32.01 \\
Intensity \cite{gao2020ProST} (Proposed,1)       & 154.09 & \textbf{7.14} & \textbf{18.90} & \textbf{6.21} & 12.85 & \textbf{23.45} \\
Intensity \cite{gao2020ProST} (Proposed,2,PE)    & 156.85 & 8.93 & 21.15 & 7.63 & 16.07 & 23.81 \\
Intensity \cite{gao2020ProST} (Proposed,2,PE,AC) & \textbf{152.98} & 7.68 & 19.06 & 6.60 & \textbf{12.21} & 24.80 \\
\hline
MICCAI \cite{gao2020ProST} (Original)          & 299.15 & 26.10 & 80.92 & 23.48 & 75.11& 65.95\\
MICCAI \cite{gao2020ProST} (Proposed,1)        & 250.42 & 8.34 & 63.48 & 6.99 & 58.41& 49.24\\
MICCAI \cite{gao2020ProST} (Proposed,2,PE)     & 241.02 & 8.56 & 58.37 & 7.29 & 53.87 & 45.29\\
MICCAI \cite{gao2020ProST} (Proposed,2,PE,AC)  & 242.99 & 8.66 & 59.94 & 7.25 & 54.00& 49.79\\
\hline
TMI \cite{gao2023ProST} (Original)          & 300.00 & 26.96 & 46.98 & 24.30 & 47.12 & 30.76\\
TMI \cite{gao2023ProST} (Proposed,1)        & 273.56 & 8.97 & 37.64 & 7.83 & 35.70 & 28.31\\
TMI \cite{gao2023ProST} (Proposed,2,PE)     & 266.65 & 9.11 & 35.29 & 7.86 & 33.33 & 27.84\\
TMI \cite{gao2023ProST} (Proposed,2,PE,AC)  & 266.26 & 9.33 & 38.21 & 8.00 & 33.89 & 35.55\\
\hline
\end{tabular}
\end{table}

\section{Experiments}
\subsection{Data and Experimental Setting}
Our X-ray simulation setup follows Gao et al. 2020 \cite{gao2020ProST} and 2023 \cite{gao2023ProST}, where they model a Siemens CIOS Fusion C-arm, featuring an image resolution of $1536 \times 1536$ with an isotropic pixel spacing of 0.194 mm per pixel and a source-to-detector distance of 1020 mm. The images are downsampled to $128 \times 128$, resulting in a pixel spacing of 2.176 mm per pixel. The source-to-iso-center distance was set at 800 mm.

We gathered twenty five CT images for our training and testing dataset from New Mexico Decedent Image Database  \cite{edgar2020NMDID}. The CT images were manually cropped to concentrate on the pelvic region and resmapled to maintain an isotropic cubic shape with 128 voxels per dimension. The pelvis anatomy was segmented automatically using Krčah et al. \cite{krvcah2011hipseg}. From the dataset, twenty scans were used for training the regression model and the registration model, and five scans were used for evaluation.

The experiments were conducted in two distinct environments. In each case, the target images $I_t$ were generated by applying a uniformly sampled pose parameters. Random rotations $(r_x,r_y,r_z)$ were applied within the ranges of $ [-20, 20] $ degrees and $ [-45, 45] $ degrees, while translations $(t_x,t_z,t_z)$ were randomly sampled within $ [-30, 30] $ mm and $ [-50, 50] $ mm along all three axes. The first environment followed the experimental range defined by Gao et al. \cite{gao2020ProST}, while the other extended this range to evaluate the robustness of the registration under larger pose variations. 

All experiments were performed on a workstation equipped with an Intel(R) Xeon(R) W-2265 CPU @ 3.50GHz (12 cores, 24 threads) and single NVIDIA RTX A4000 GPUs with 16 GB of VRAM. The system runs Ubuntu 22.04.4 LTS and the models were trained and evaluated using the CUDA 13.0 toolkit. The software environment was managed using conda and included Python 3.10.15.


\subsection{Models, Training, and Evaluation}
Three different optimization-based pose estimators were implemented and tested with and without the proposed initialization. As these estimators are introduced by the same authors and reuse components of each other, we provide specific names to three methods instead of using their publication names/years, which otherwise would induce collisions. The first model \textbf{Intensity} is the intensity-based registration method with ProST \cite{gao2020ProST} (the projection model) using \cite{penney1998gradncc} without any additional embedding or learned similarity. Effectively, it is differentiable DRR with image similarity, and contains no learnable parameters. The second model \textbf{MICCAI} is the embedding-based registration introduced in the same Gao et al. \cite{gao2020ProST}, which uses  ProST, but registration is conducted in the learned embedding space. The third model \textbf{TMI} employs a different learned similarity scheme, as introduced by Gao et al. 2023 \cite{gao2023ProST}. For all the models, registration was conducted until convergence, and stopped at 300 iterations if not converged. 


The three previously described initialization options \{\textbf{(Proposed,1)}, \textbf{(Proposed,2,PE)}, \textbf{(Proposed,2,PE,AC)}\} were trained using 8,800 generated samples from the training volumes. Each of these was used to generate starting points for 2,200 test cases, alongside the naive \textbf{(Original)} initialization. The discrepancy between the ground truth pose and method results from each of the three pose-estimators for each of the four initialzation options were then measured, using root mean squared error (RMSE) and mean absolute error (MAE), separated by rotation parameters and translation parameters. Additionally, the \textbf{Initializer} $\theta_0$ discrepencies without the pose-estimator optimizations were also measured as reference values.
In the evaluation of translation error, we present separate results for in-plane and out-of-plane translations. In-plane translation refers to displacement along the $x$- and $y$-axes, while out-of-plane translation corresponds to displacement along the $z$-axis. Unlike in-plane translation, which generally achieves accurate registration, out-of-plane translation is more challenging due to the inherent depth ambiguities in 2D/3D registration. Therefore, separate columns are reported to highlight the distinct performance characteristics of each translation.

\section{Results}

\label{sec:results}

\begin{figure}[t]
    \centering
    \begin{subfigure}{0.90\textwidth}
        \includegraphics[width=\textwidth]{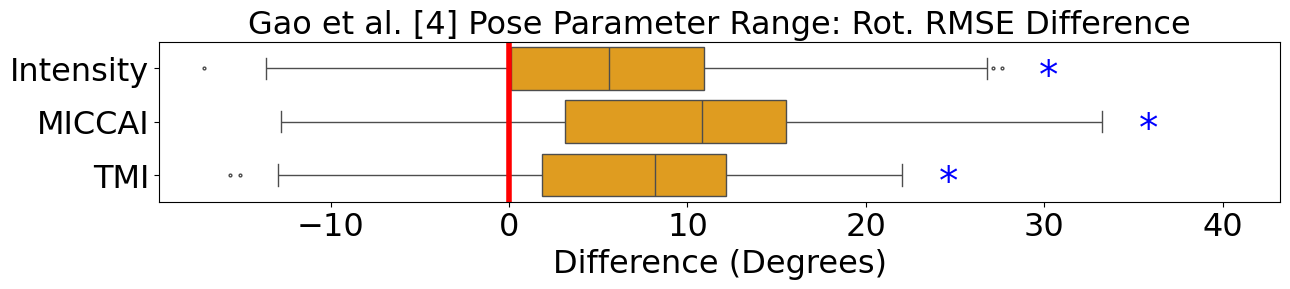}
    \end{subfigure} \\
    \begin{subfigure}{0.90\textwidth}
        \includegraphics[width=\textwidth]{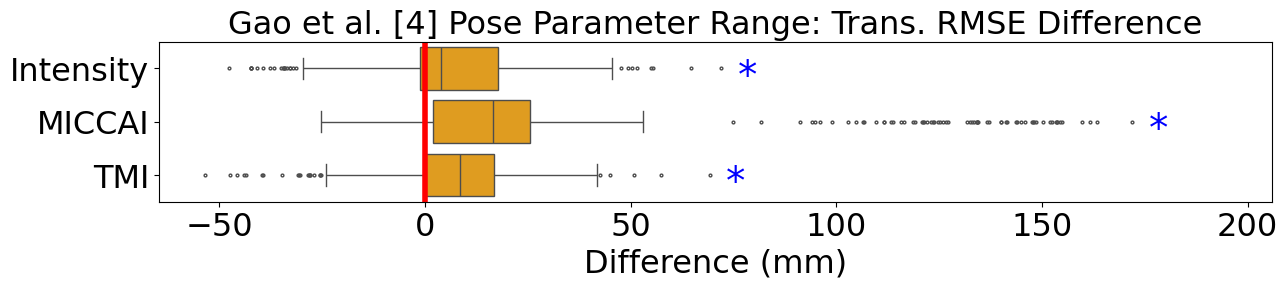}
    \end{subfigure} \\
    \begin{subfigure}{0.90\textwidth}
        \includegraphics[width=\textwidth]{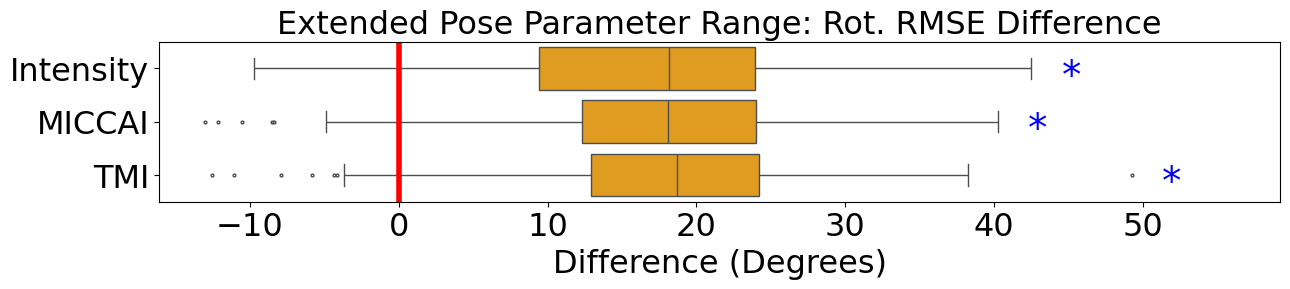}
    \end{subfigure}\\
    \begin{subfigure}{0.90\textwidth}
        \includegraphics[width=\textwidth]{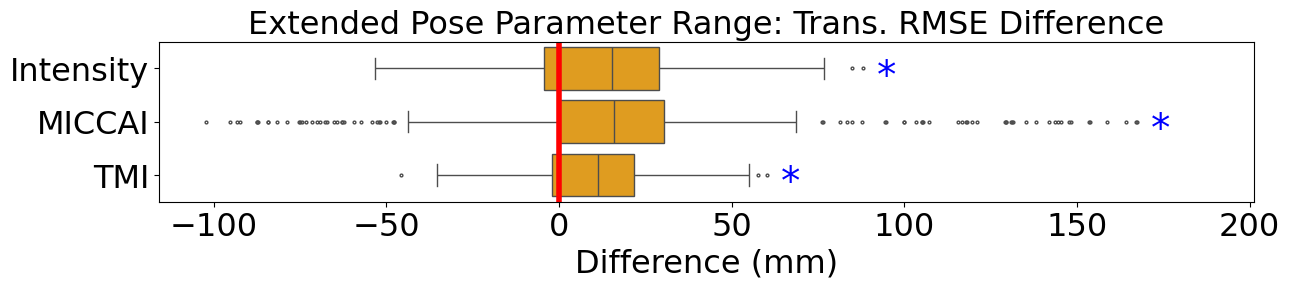}
    \end{subfigure}

    \caption{Box plots of the difference in RMSE between the (Original) and (Proposed,1) methods for both rotation and translation under different experimental conditions. Statistical significance is marked with based on single tail t-tests, * denotes $p < 0.0001$.}
    \label{fig:box plot with t-test}
\end{figure}

As shown in Table~\ref{tab:average}, overall Proposed initializations achieve improved registration accuracy as measured by RMSE and MAE compared to the Original initializations across both experimental settings for all pose estimators. While the difficult setting presents greater challenges due to larger pose variations, the proposed methods still demonstrate better performance. However, out-of-plane translation remains challenging to optimize compared to in-plane translation.

The initializer itself has surprisingly high accuracy, but does not out-perform the optimization based pose-estimators when combined with the proposed initializer. This fits our intuition that the initializer can produce ``coarse but close-enough'' estimates, after which the optimizer can refine those estimates.

The box plots in Fig.~\ref{fig:box plot with t-test} illustrate the difference in RMSE between the Original and Proposed methods for both rotation and translation. The positive values indicate improved accuracy and the red vertical line at zero serves as a reference to distinguish improvements from deterioration. The results confirm that the proposed methods consistently reduce RMSE across different experimental condition. Single tail t-tests indicate significant differences (above zero), confirming that the observed improvements are unlikely to be due to random chance. 

\begin{figure}[t]
    \centering
    \includegraphics[width=0.98\textwidth]{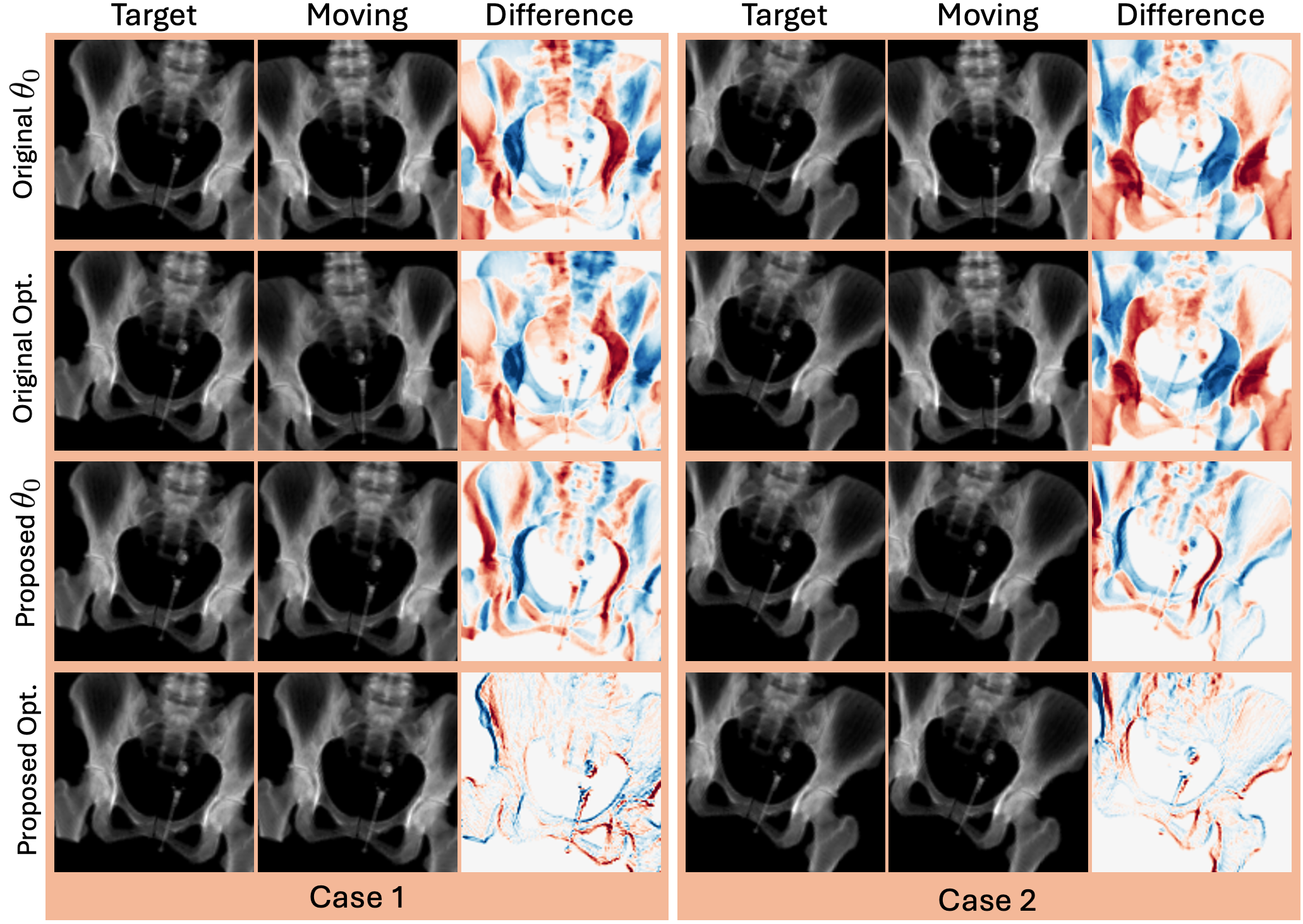}
    \caption{Comparison of initialization and registration results using ProST on 2 sample cases using the original (naive) initialization and the proposed initialization. The third column shows difference maps (moving minus target, red:positive, blue:negative).} 
    \label{fig:difference}
\end{figure}

Fig.~\ref{fig:difference} provides a comparison of registration accuracy between the Original and Proposed initialization. As shown in second and the fourth row, Original fails to register to the target, but the Proposed demonstrates better alignment, exhibiting fewer intensity differences from the target. Notably, as shown in the third row, the predicted pose $(\theta_0)$ allows the registration to start closer from the ground truth.
\section{Discussion}
\label{sec:discussion}

Our work demonstrates that a simple, regression-based initialization network can effectively overcome the initialization sensitivity that is a known limitation of many optimization-based 2D/3D pelvis registration methods. This hybrid approach provides a "close-enough" starting point that is sufficient to guide the optimizer away from local minima, resulting in a consistent and significant improvement in registration accuracy. Furthermore, this enhanced initialization drastically reduces the number of iterations required for convergence, improving the computational efficiency.

Despite these promising results, this study has several important limitations. First, while our method improved all pose parameters, the out-of-plane (z-axis) translation remains the most challenging to optimize. This reflects an inherent depth ambiguity in single-view 2D/3D registration that our initializer, while helpful, does not fully solve. Second, our model was trained and validated on a limited dataset of 25 CT volumes from a single database. The dataset size was constrained by the time-consuming manual preprocessing required for each scan, which means its generalizability to diverse datasets remains unconfirmed. Finally, our preprocessing pipeline was not fully automated, as it relied on manual cropping of the pelvic region. Future work must therefore focus on validating this framework on large, multi-center clinical datasets of real fluoroscopy images and investigating methods to better resolve the persistent depth ambiguity.

\section{Conclusion}

This work introduced an improved 2D/3D registration framework that enhances pelvic pose estimation by incorporating a regression-based initialization. This approach addresses the limitations of traditional methods that often suffer from slow convergence and poor initialization sensitivity. By predicting an initial pose estimate,  we reduce the number of registration iterations while achieving higher registration accuracy in both rotational and translational parameters. Experimental results demonstrated consistent improvements over conventional methods across different experimental condition. These findings show the importance of data-driven initialization strategies enhance registration efficiency and robustness.

\backmatter







%

\section*{Statements and Declarations}
\begin{itemize}
\item Funding: This work was supported in part by NSF 2321684 and a VISE Seed Grant
\item Conflict of interest: The authors have no competing interests to declare that are relevant to the content of this article
\item Ethics approval and consent to participate: Not applicable
\item Consent for publication: Not applicable
\item Data availability: This study was conducted using data from the New Mexico Decedent Image Database \cite{edgar2020NMDID}, the Free Access Decedent Database funded by the National Institute of Justice grant number 2016-DN-BX-0144.
\item Materials availability: Not applicable
\item Code availability: Link to Github repository provided in the abstract
\item Author contribution: Yehyun Suh: Conceptualization, Methodology, Software, Validation, Formal Analysis, Investigation, Data Curation, Visualization, Writing – Original Draft, Writing – Review and Editing; J. Ryan Martin: Methodology, Validation; Daniel Moyer: Conceptualization, Methodology, Validation, Resources, Writing – Review and Editing, Supervision, Project Administration, Funding Acquisition
\end{itemize}

\bibliography{ref}

\end{document}